\pdfoutput=1

\documentclass[11pt]{article}

\usepackage{latex/acl}

\usepackage{times}
\usepackage{latexsym}

\usepackage[T1]{fontenc}

\usepackage[utf8]{inputenc}

\usepackage{microtype}

\usepackage{inconsolata}

\usepackage{comment}
\usepackage{amsmath}
\usepackage{amssymb}
\usepackage{graphicx}
\usepackage{tabularx}
\definecolor{darkgreen}{RGB}{44, 143, 0}
\title{Towards 
Reducing Diagnostic Errors with Interpretable Risk Prediction}

\author{Denis Jered McInerney \\
  Northeastern University \\
  \texttt{mcinerney.de@northeastern.edu} \\\And
  William Dickinson \\
  Brigham and Women's Hospital \\
  \texttt{wdickinson@bwh.harvard.edu} \\\AND
  Lucy C. Flynn \\
  Brigham and Women's Hospital \\
  \texttt{lcflynn@mgb.org} \\\And
  Andrea C. Young \\
  Brigham and Women's Hospital \\
  \texttt{acyoung@bwh.harvard.edu} \\\AND
  Geoffrey S. Young \\
  Brigham and Women's Hospital \\
  \texttt{gsyoung@bwh.harvard.edu} \\\And
  Jan-Willem van de Meent \\
  University of Amsterdam \\
  \texttt{j.w.vandemeent@uva.nl} \\\And
  Byron C. Wallace \\
  Northeastern University \\
  \texttt{b.wallace@northeastern.edu} \\}

\begin{document}
\maketitle
\begin{abstract}
    Many diagnostic errors occur because 
    clinicians cannot easily access 
    relevant information in patient Electronic Health Records (EHRs). 
    In this work 
    we propose a method to use LLMs to identify 
    pieces of evidence in 
    patient EHR data that 
    indicate increased or decreased risk of specific diagnoses; our ultimate aim is to increase access to evidence and reduce diagnostic errors.
    In particular, we propose a Neural Additive Model to make predictions backed 
    by evidence with individualized risk estimates at time-points where clinicians are still uncertain, aiming to specifically mitigate delays in diagnosis and errors stemming from an incomplete differential.
    To train such a model, it is necessary to infer temporally fine-grained 
    retrospective labels of eventual ``true'' diagnoses. 
    We do so with LLMs, to ensure that the input text is from \textit{before} a confident diagnosis can be made.
    We use an LLM to retrieve an initial pool of evidence, but then refine this set of evidence according to correlations learned by the model.
    We conduct an in-depth evaluation of the usefulness of our approach by simulating how it might be used by a clinician 
    to decide between a pre-defined list of differential diagnoses.\footnote{We make our code publicly available for: 1) retrieving evidence and target diagnoses from EHR text in the form of a gym environment---\url{https://github.com/dmcinerney/ehr-diagnosis-env}, 2) training agents---\url{https://github.com/dmcinerney/ehr-diagnosis-agent}, and 3) visualizing and annotating predictions---\url{https://github.com/dmcinerney/ehr-diagnosis-env-interface}.}
\end{abstract}

\begin{figure}
    \centering
    \includegraphics[scale=.106]{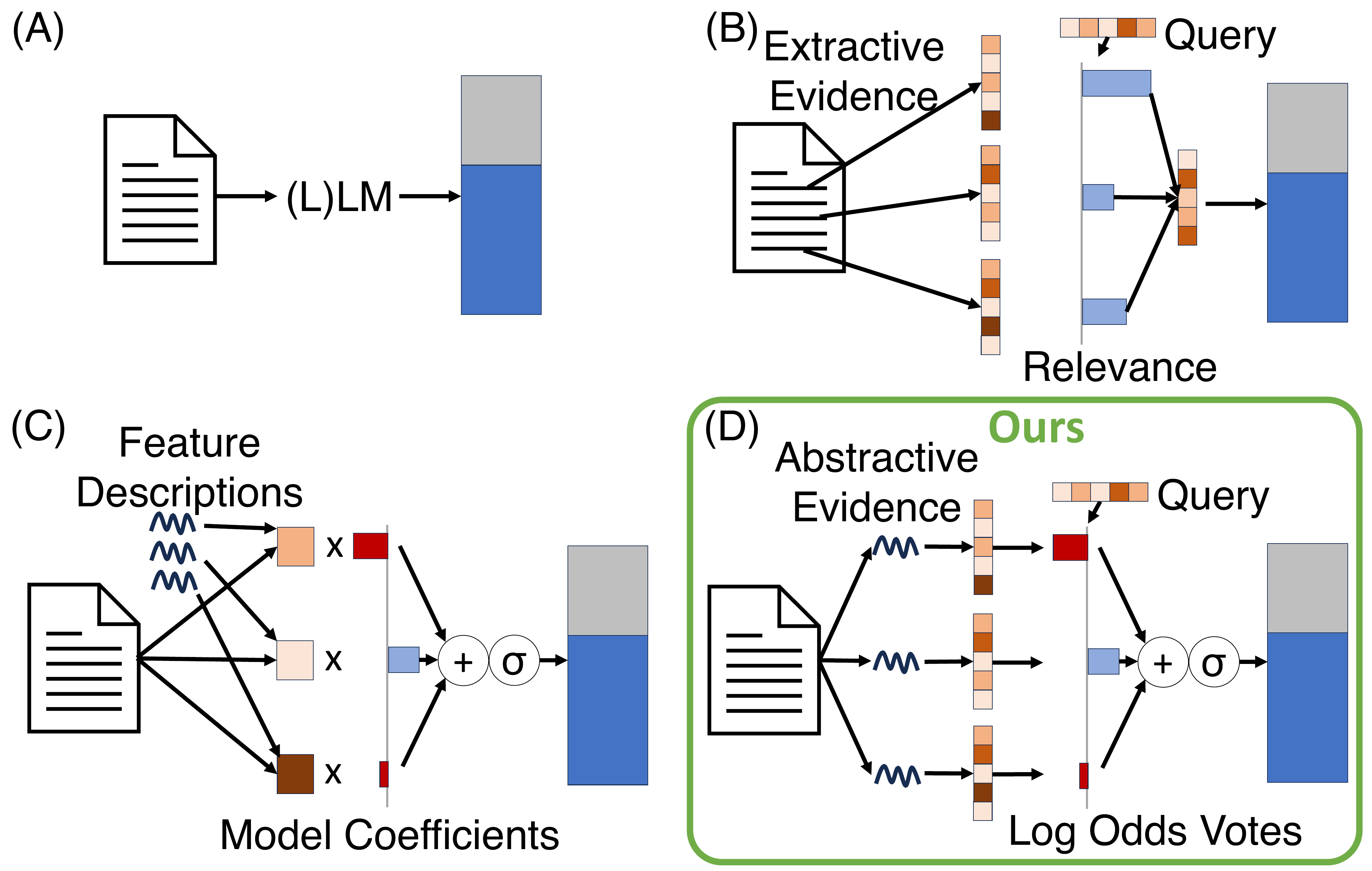}
    \caption{\textbf{Inherently ``interpretable'' approaches to prediction}.
    Typically, `interpretable' models trade off between the expressiveness of intermediate representations and the faithfulness of the resulting interpretability to the models' true mechanisms.
    Our approach (D) manages to use very expressive intermediate representations in the form of abstractive natural language evidence while still maintaining true transparency during aggregation of this evidence.
    See Table \ref{tab:inherently-interpretable-approaches} for more details.}
    \label{fig:inherently-interpretable-approaches}
\end{figure}

\begin{figure*}
    \centering
    \includegraphics[scale=.5]{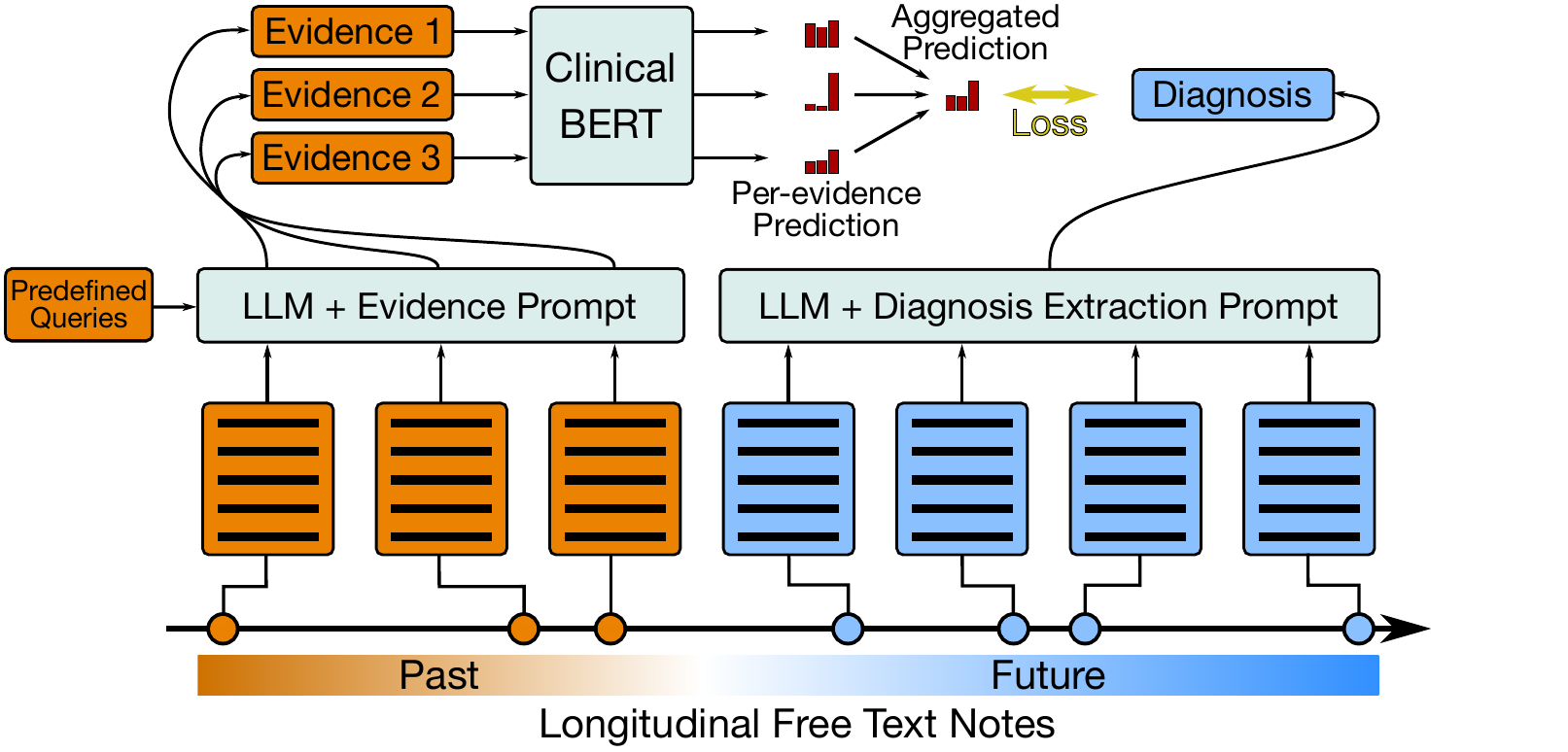}
    \caption{\textbf{Explainable Risk Prediction and Training.} An overview of our approach. Left: We retrieve evidence snippets from past notes with an LLM for predefined queries posed by a clinician. Then we use our risk prediction model to estimate risk of various diagnoses given each piece of evidence individually, and aggregate these scores. Right: We automatically extract diagnosis `labels' from future reports with an LLM to use to train the risk predictor.}
    \label{fig:concept}
\end{figure*}

\section{Introduction}



A major source of poor patient outcomes and unnecessary costs in healthcare are missed or delayed diagnoses.
A recent report estimated that diagnostic errors result in around 795,000 serious harms annually \cite{Newman-Tokerbmjqs-2021-014130}.
Furthermore, many diagnostic errors result from information transfer problems \cite{10.1001/archinternmed.2010.146}.
This is unsurprising given ``note bloat'', i.e., the widespread problem of information overload in EHR notes, often due to copied or irrelevant information which obfuscates relevant information.
All of this motivates the potential of providing more efficient mechanisms to access relevant information in EHRs as a means to reduce these errors.

% ...

One approach to helping practitioners make use of EHR is to train NLP models
to provide predictions about patient risk for various illnesses \cite{rasmy_med-bert_2021, Li2021HiBEHRTHT, Yang2023TransformEHRTE}, but these systems are often lack transparency.
Even when systems have high accuracy, clinicians may still prefer simple linear models as clinical decision support tools \cite{10.1093/jamia/ocw042}.
Prior work has focused on developing inherently interpretable\footnote{Interpretability is a famously ambiguous term; we are focused on having explicit measure of the contribution of individual pieces of evidence to an output.} models with minimal tradeoff in predictive performance, e.g., in the general domain with Neural Additive Models \cite{Agarwal2020NeuralAM} and in healthcare with $\textrm{GA}^2$Ms \cite{caruana_intelligible_2015}.
Recently, zero-shot instruction-tuned LLMs have been shown capable of extracting information from clinical text \cite{agrawal-etal-2022-large}, which in turn facilitates interpretable predictions \cite{mcinerney-etal-2023-chill, Alsentzer2023ZeroshotIP}.

In this work, we combine the power and flexibility of zero-shot instruction-tuned LLMs with the transparency and modeling ability of Neural Additive Models (NAMs) to train a risk-prediction model that can also surface evidence to support predictions.
We use an LLM (FLAN-T5-XXL; \citealt{chung2022scaling}) to generate abstractive ``evidence'' from EHR, which is then processed by a simpler model (Clinical BERT; \citealt{alsentzer-etal-2019-publicly}) to produce features for a Neural Additive Model (Figure~\ref{fig:concept}). 
This provides flexibility---the model can make inferences and condense information into fluent text snippets---but brings risk of ``hallucinations''. 

This approach is ``interpretable'' insofar as it produces ``evidence'' in the form of human-understandable intermediate variables: Abstractive text with associated risks, providing insight into factors that informed predictions. 
Related approaches to ``interpretability'' (Figure \ref{fig:inherently-interpretable-approaches}) include using relevance scores to weight and combine information from different sentences (B), and those that use LLM prompts to infer feature values (C).
Our approach permits greater flexibility than (C), while maintaining a more faithful interpretability in comparison to (B); see Table \ref{tab:inherently-interpretable-approaches}. 

One complication 
is that we would like 
fine-grained, accurate labels to train our predictor (see section \ref{sec:synthetic-label-extraction}); ICD codes do not meet these criteria \cite{searle2020experimental}.  
Instead of ICD codes, which are noisy and temporally coarse (observed at the end of an encounter with discharge summaries), we propose to synthetically extract diagnosis labels from each report using an LLM.
In some cases, this has been shown to be more aligned with true diagnoses \cite{Alsentzer2023ZeroshotIP}.

We focus our evaluation on how this system impacts clinical decision-making.
Specifically, we examine settings where risk of misdiagnosis is high and the consequences severe.
Our methods work within the confines of data present in electronic health record, which allows the model to be trained on any EHR.
LLMs can be run locally and are only used for inference, so privacy and compute resources are not an issue.

Our contributions are summarized as follows:

    \vspace{0.3em}
\noindent \textbf{Interpretable Risk Prediction with LLMs.}
    We propose an approach to risk prediction that offers a particular form of interpretability in that it can expose faithful relationships between specific pieces of retrieved evidence and an output prediction.

  \vspace{0.3em}
\noindent \textbf{Extracting Future Targets with LLMs.}
    We present a method to extract target diagnoses for use in training from the unstructured text in the future of a patient's medical record that are more granular than ICD codes in the time dimension, and we validate with clinician annotations that the extracted labels are accurate.

   \vspace{0.3em}
\noindent \textbf{In-depth Annotation of Usefulness.} We validate how much evidence-wise interpretability can positively impact a clinician's expert judgement in high-impact settings which feature the greatest risk of misdiagnosis.



\begin{table*}[]
    \centering
    \footnotesize
    \begin{tabular}{p{3.5cm} p{3.9cm} p{3.5cm} p{3.5cm}}
    \hline
        Modeling Approach & Intermediate representation(s) & Aggregation & Interpretability \\
        \hline
        \vspace{-.5em}{\color{red}} (A) Direct \textbf{Black-box} Prediction (e.g., zero/few-shot, fine-tuned LLM) & \vspace{-.5em}{\color{red} None} & \vspace{-.5em}{\color{red} CLS or last token embedding + classification or LM head} & \vspace{-.5em}{\color{red} No inherent interpretability} \\
        \vspace{-.5em}{\color{red}} (B) Aggregating chunked input with \textbf{relevance weights} & \vspace{-.5em}{\color{red} \textit{Extractive}} {\color{darkgreen} text snippets} &  \vspace{-.5em}{\color{red} Weighted avg.~of CLS embeddings + class.~head} & \vspace{-.5em}{\color{darkgreen} Positive, real-valued relevance scores per query} \\
        \vspace{-.5em}{\color{red}} (C) \textbf{Logistic regression} with LLM-inferred features & \vspace{-.5em}{\color{darkgreen} \textit{Inferred}}, {\color{red} real-valued numbers} relating to  predefined natural language queries & \vspace{-.5em}{\color{darkgreen} Logistic regression} & \vspace{-.5em}{\color{darkgreen} Negative and positive real-valued static model coefficients} \\
        \vspace{-.5em}{\color{red}} (D) \textbf{Log odds voting} with LLM-inferred text snippets (ours) & \vspace{-.5em}{\color{darkgreen} \textit{Inferred}/\textit{abstractive} text snippets} relating to predefined natural language queries & \vspace{-.5em}{\color{darkgreen} Neural Additive Model (conditioned on the query/condition vector)} & \vspace{-.5em}{\color{darkgreen} Negative and positive real-valued dynamic impact scores} \\
        \hline
    \end{tabular}
    \caption{Types of interpretability afforded by the different modeling approaches for EHR data visualized in Figure \ref{fig:inherently-interpretable-approaches}. Red and green denote negative and positive aspects of each model.}
    \label{tab:inherently-interpretable-approaches}
\end{table*}

\section{Dataset}

We use MIMIC-III \cite{johnson2016physionet, johnson_mimic-iii_2016}, an open-source dataset of EHRs from ICU patients. The ICU is one of the hospital settings (along with, e.g., the ER and Radiology) where misdiagnosis or delayed diagnosis are often caused by incomplete information, since clinicians typically do not have enough time to fully examine a patient's EHR.

In healthcare, cancer, infection, and vascular dysfunction (termed the ``big three'') account for about 75\% of all mis-diagnosis-related harms \cite{Newman-Tokerbmjqs-2021-014130}.
Within the ICU, the latter two categories mostly manifest as pneumonia, and pulmonary edema (which in this paper we treat as interchangeable with congestive heart failure). For this reason, we will focus on predicting the risk of ICU patients for cancer, pneumonia, and pulmonary edema.
These are also conditions for which clinical correlation with notes from the past EHR is important for diagnosis.
We use all patients in the MIMIC dataset so that we have both negative and positive examples of the conditions.
We include additional details regarding the dataset and preprocessing in appendix section \ref{sec:dataset-and-preprocessing}.



\section{An Interpretable Risk Prediction Model} \label{sec:methods}



We propose a multi-stage approach to risk prediction, capitalizing on a modern LLM, FLAN-T5-XXL \citep{chung2022scaling, 51119} in this case, to implement each of the following steps. 
   
\vspace{0.3em}
\noindent {\bf Retrieval (Section \ref{sec:evidence-retrieval}).} We generate abstractive evidence from free text notes by prompting an LLM with appropriate queries. The evidence snippets provide a form of interpretability, in that they can be inspected directly to verify predictions. 
    
\vspace{0.3em}
\noindent {\bf Risk Prediction (Section \ref{sec:risk-prediction}).} We input the evidence into the risk predictor, which models relationships between the evidence and each of the potential diagnoses and outputs multi-label classification probabilities, i.e. the predicted risk that the patient will be diagnosed with each condition.

\vspace{0.3em}
\noindent {\bf Evidence Re-ranking (Section \ref{sec:evidence-reranking}).} The retrieved evidence may still be too large a pool  to review given the time constraints of the clinician.
    Therefore, we re-rank the evidence so as to only show that which promotes risk predictions that most deviate from the baseline risks of each condition.

\vspace{0.3em}
To train risk prediction models we use use synthetic labels extracted from \emph{future} notes in a patient's record (Section \ref{sec:diagnosis-extraction}).
Figure \ref{fig:concept} provides an overview of our model and training approach.



\subsection{Evidence Retrieval}\label{sec:evidence-retrieval}


Following prior work \cite{ahsan2023retrieving}, we use 
a sequential prompting strategy to retrieve evidence that is relevant to a queried diagnosis or a risk factor.
Specifically, we first ask the LLM for a binary response as to whether evidence for a condition exists; if the answer is affirmative, 
we then issue a second prompt tasking the LLM to generate supporting evidence. 
Formally, we define the evidence retrieved for report $n$ and query $q_i$ as follows:
\begin{equation}
    e_{n, q_i} = 
    \begin{cases}
        \textrm{GetEvidence}(r_n, q_i) \\
        \quad \textrm{if\ EvidenceExists}(r_n, q_i)=\textrm{``yes''} \\
        \texttt{null} \quad \textrm{otherwise} \\
    \end{cases}
\end{equation}
where ``$\mathrm{GetEvidence}$'' and ``$\mathrm{EvidenceExists}$'' represent the corresponding prompt functions.

This approach does have limitations.
For example, it cannot produce more than one snippet of evidence per report/query pair.
Retrieved evidence may also be abstractive rather than extractive, which introduces the risk of model ``hallucinations'', but permits flexibility and interpretability \cite{ahsan2023retrieving}.
It also significantly reduces the amount of text (therefore requiring a relatively small context window) by going from all reports to sentence-length snippets for some reports.
The resulting ``summarization'' in the form of evidence snippets is also controllable through the querying process and works zero-shot, i.e., it requires no specialized or in-domain training.
Queries, in the form of the 3 diagnoses considered and risk factors written by a clinician co-author, are shown in appendix Table \ref{tab:risk-factors}.
We present further details regarding the evidence retrieval prompts in Appendix \ref{sec:evidence-prompt}.

\subsection{Risk Prediction}\label{sec:risk-prediction}

Because a patient can have more than one diagnosis, we treat risk prediction as a multi-label classification problem where each label corresponds to a diagnosis. 
To realize interpretability, we 
use a Neural Additive Model \cite{Agarwal2020NeuralAM}.
Specifically, we do not model \emph{interactions} between 
evidence snippets.
Instead, we predict scores individually for each piece of evidence, and average these\footnote{Neural Additive Models typically use a sum instead of an average, but we found that given varying amount of evidence retrieved, it worked better to use an average.} to obtain a logit for risk prediction:
\begin{equation}\label{eq:likelihood}
    p(\hat y_{i} = 1 | e_{1:E}) = \sigma(b_i + w_i \cdot (\frac{1}{E}\sum_{j=1}^E f^\mathrm{BERT}_\theta(e_j)))
\end{equation}
where $w_i \in \mathbb{R}^d$ is the embedding of diagnosis $i$, $e_{1:E}$ is the flattened list of evidence snippets\footnote{We add the query term used to retrieve the evidence and relative date of the evidence before serving it as input, which we describe in greater detail in Appendix \ref{sec:risk-prediction-inputs}. Also note that we use evidence surfaced by all queries for all predictions.} with $\texttt{null}$ evidence omitted, $f^\mathrm{BERT}_\theta$ is the ClinicalBERT \cite{alsentzer-etal-2019-publicly} $\texttt{[CLS]}$ embedding function (which yields a $d$-dimensional vector), and $b_i \in \mathbb{R}$ is the bias for diagnosis $i$.
The prior over conditions can be defined as the same equation excluding the evidence term: $p(\hat y_i) = \sigma(b_i)$, and the \textbf{relative risk} follows as $p(\hat y_i | e_{1:E}) / p(\hat y_i)$.

While the bias could be learned, 
we instead simply set it to the inverse sigmoid of the observed prevalence of the disease in the training sample distribution: $b_i = \sigma^{-1}(\mathrm{prevalence}^\mathrm{train}_i)$.
This means that if we wanted to transfer the model to a new population, where the prevalence differed but the contributions of different evidence were assumed to remain, we could simply update the $b_i$ term.

Excluding interactions between evidence snippets is a sacrifice in model complexity, but it also allows us to compute an interpretable ``vote'' for any individual piece of evidence as
\begin{equation}\label{eq:individual_evidence_likelihood}
    p(\hat y_{i} | e_j) = \sigma(b_i + w_i \cdot f^\mathrm{BERT}_\theta(e_j))
\end{equation}
and compute an individualized relative risk for each piece of evidence using this value.

Conveniently, forcing the bias term to be the inverse sigmoid of the training prevalence, by definition, also means we can interpret the evidence term in Equations \ref{eq:likelihood} and \ref{eq:individual_evidence_likelihood} as the \textbf{log odds ratio}, i.e., the difference between the logits when conditioning vs. not conditioning on the evidence.
The model is effectively estimating this log odds ratio directly.
This variable's expected value does not change if we sample conditions for training with a frequency different from the the natural prevalence of the conditions \cite{simon2001understanding}.
Because of this, we can estimate the likelihood and the relative risk during inference on a differently sampled population by simply changing the bias term in the prior and in equations \ref{eq:likelihood} and \ref{eq:individual_evidence_likelihood} to reflect the estimate of the natural prevalence of the conditions \citep{zhang1998s}, which we can get from the training set before sampling: $b_i' = \sigma^{-1}(\mathrm{prevalence}^\mathrm{train}_i)$.

\subsection{Evidence Re-ranking}\label{sec:evidence-reranking}

Because of the simplicity of the risk prediction, we can use the internal variables it exposes to re-rank evidence.
The intuition behind the re-ranking is that the most important evidence will be that which most changes our risk assessment from the prior over the diagnoses, and we would like the chosen metric to capture this across all of the potential diagnoses.
We use Mean Squared Error (MSE) of the predicted logits with the logits of the prior $p(y)$.
This makes the formulation of the MSE metric simple as the mean (over $Q$ conditions) of the squares of the log odds ratio for a piece of evidence:
\begin{equation}\label{eq:mse}
\begin{split}
    \mathrm{MSE}(\sigma^{-1} p(\hat y | e_j), \sigma^{-1} p(\hat y)) = \\
    \frac{1}{Q}\sum_{i=1}^Q ( w_i \cdot f^\mathrm{BERT}_\theta(e_j))^2.
\end{split}
\end{equation}
It is necessary to use the \textit{log odds} ratio term in this score function because we care not only about increasing but also about decreasing the probability of a condition, so it makes most sense to compare and sum these two different effects in log space.
The reason to choose MSE over other scores (e.g. the absolute distance) comes from the intuition that it is more important to see the evidence that is ``very opinionated'' about one condition rather than to see evidence that is ``slightly opinionated'' about many.
Therefore, it is necessary to square this log odds ratio before averaging across conditions to reflect this idea when sorting evidence.

\section{Certain Diagnosis Extraction}\label{sec:diagnosis-extraction}

We make an assumption about the EHR of patients that eventually receive a diagnosis that there is some period of time in the record where a diagnosis is ``uncertain'' before it becomes ``certain'', and the eventual ``certain'' diagnosis is correct.
Of course just because a diagnosis is definitive as noted by clinician in the record does not necessarily mean that it is correct---sometimes clinicians are wrong.

However, it is hard to detect such cases, so here we focus on reducing delayed diagnosis errors where we assume some evidence in the medical record from that ``uncertain'' period could have influenced a clinician to make a diagnosis or order a certain kind of test sooner than they did, or keep a diagnosis in the running list of differentials for longer.
If notes are incorporated into the input where the diagnosis is already certain, the prediction problem becomes too easy, which is why a time-wise fine-grained label is necessary---such a label could more accurately weed out all of this obvious evidence.
To extract these certain diagnoses with an LLM, we use three sequential prompts and a normalization step.

\subsection{3-Stage Extraction with LLMs}\label{sec:synthetic-label-extraction}

In this section we describe the prompts for certain diagnosis extraction, which are shown in full in Appendix \ref{sec:full-prompts}.
Following prior work \cite{ahsan2023retrieving}, we first prompt the LLM with a binary question asking if there exists a confident diagnosis for a patient.
If the answer is ``yes'', we then ask the model for the diagnoses.
Unfortunately, creating a list of diagnosis terms from the answer to this prompt is not just a matter of parsing because we found that the model will often return extended phrases that are not easily mapped to diagnoses.
Therefore, we issue one more prompt that only takes in the output of the previous prompt to create a structured list of diagnostic terms.
We then parse this final output of the LLM into a list of strings.

\subsection{Normalization}

To normalize produced diagnostic terms, we take a two-step apporach.
First we use string matching heuristics to handle easy cases.
Then we embed sentences with {\tt SentenceTransformers} (\citealt{wang2020minilm, reimers-2019-sentence-bert}; specifically, \texttt{all-MiniLM-L6-v2}) and calculate cosine similarities, matching a term in the parsed list to the most similar term (with similarity $>$.85) in the predefined set (``cancer'', ``pneumonia'', and ``pulmonary edema'').
We ignore terms with no match. 


\section{Evaluation}\label{sec:eval}

Because our targets are synthetically generated using an LM, we first evaluate how well our labels align with the ``ground truth'' (Section \ref{sec:eval-targets}).
Next, we aim to evaluate how well the model can realistically help with risk prediction.
Though it is straightforward to assess the accuracy of the risk prediction itself---we use the standard metrics of precision, recall, F1 and AUROC scores to compare to various uninterpretable baselines---it is not as easy to assess what we really care about: How helpful is the interpretability offered by the proposed model to clinicians (section \ref{sec:eval-predictions})? 
For this we resort to manual evaluation by our clinical co-authors and develop bespoke interfaces to facilitate annotation.

\subsection{Future Target Extraction}\label{sec:eval-targets}

To evaluate how well the LLM extracts targets in the form of ``confident'' diagnoses, we enlist our clinical collaborators to annotate the precision with which the LLM infers ``confident'' diagnoses.
In particular, for every report where one of the three diagnoses---cancer, pneumonia, and pulmonary edema---was automatically extracted, an ICU clinician is first tasked with answering the question ``Is [diagnosis] a confident diagnosis of the patient according to the report?''. 
If the answer is ``yes'', they are asked: ``Is it likely that this confident diagnosis could be identified in earlier reports?''.

\subsection{Risk Prediction Interpretability}\label{sec:eval-predictions}

To assess the viability of clinicians using this model in practice, we collect in-depth annotations intended to simulate the real-world use of this technology.
We evaluate a number of baseline models and model ablations to 
assess the relative benefits of different model components. 

\paragraph{Interface and Annotations}

To conduct annotations, we develop an interface that simulates as closely as possible the envisioned use case: A clinician is seeing an ICU patient's chart for the first time and trying diagnose the patient or determine what they are at risk of. 
The clinician may not have much time to spend with the patient's chart, so we ask clinician annotators to work quickly---specifically, to try and keep annotation time to a few minutes---and we record the amount of time they take to review the patient's record.
When they are done, the annotation process starts, and though they are allowed to access the patient's notes, they are encouraged not to.

We first ask if a diagnosis is noted explicitly in the patient's record.
Given that we are aiming to evaluated records where the diagnosis is not yet clear, we skip the rest of the annotations on the instance if a diagnosis is explicit.
If not, we ask for estimates of the likelihood (``unlikely'', ``somewhat likely'', or ``very likely'') of each of the possible conditions.
Note that we explicitly do not show any model predictions until after this question, to avoid bias.
Then, we show the annotator the model predictions and ask if the predicted risk for the conditions aligns with intuition.

Moving onto the evidence (appendix Figure \ref{fig:evidence_ann_interface}), we allow the annotator to look at the sorted evidence one snippet at a time along with the individualized risk prediction only based on that snippet.
The annotator notes the usefulness of the evidence with respect to each condition. 
If the evidence is useful, they are asked whether or not the impact of this evidence on the risk scoring (for the particular condition) aligns with intuition, and whether the annotator remembers seeing this piece of evidence during their initial review of the patient's notes.
After two pieces of evidence, if the annotator feels like more evidence is needed to form a reasonable opinion of the patient's risk, they can request more evidence snippets (up to a maximum of 10), annotating each as they go.
Finally, the annotator is asked if any of the evidence presented impacted their original assessment of likelihood.

\paragraph{Ablations}
\label{section:ablations}

While the task of risk prediction is standard, there is less work on the task of surfacing relevant evidence (abstracted or extracted) to support such predictions. 
Consequently, there is not a large set of baselines to serve as natural comparators to our approach. 
Therefore, in our analysis we focus on 
showing the importance of each component of our model through ablations.
We can decompose our approach into two evidence retrieval components, generating the evidence, which we refer to as ``\textbf{LLM Evidence}'' and reranking it, which we refer to as ``\textbf{Log Odds Sorting}''.
The following ablations show the importance of both of these components in identifying useful evidence.

We use prior work \cite{ahsan2023retrieving} as a starting point for generating the evidence, so it is natural to ask what that component can do by itself without re-ranking using the risk prediction scores for each piece of evidence.
A natural comparison is to present the same evidence retrieved but in a \emph{random} or \emph{reverse chronological} order (as recency is probably important). 
But we can also use the model certainty in evidence, given that this has been shown to correlate with the utility of snippets \cite{ahsan2023retrieving}.
We adopt this approach for comparison and call it 
``\textbf{Confidence Sorting}''.

It is also natural to question the importance of using the language model to abstractively generate evidence at all.
We might instead simply use every sentence in the report as evidence and train our prediction model with this retrieved evidence, re-ranking it in the normal way (``Log Odds Sorting'') with the prediction model's scores.
We call this the ``\textbf{All EHR}'' model.

\section{Results and Discussion}
\label{section:results}

\begin{table*}[]
    \centering
    \footnotesize
    \resizebox{\textwidth}{!}{
    \begin{tabular}{c|c c c c|c c c c|c c c c}
& \multicolumn{4}{c|}{LLM Evidence+Confidence Sorting} & \multicolumn{4}{c|}{Raw EHR+Log Odds Sorting} & \multicolumn{4}{c}{LLM Evidence+Log Odds Sorting} \\
Annotator & Inst. & Evid. & Rep. & Percent Useful & Inst. & Evid. & Rep. & Percent Useful & Inst. & Evid. & Rep. & Percent Useful \\
\hline
1 & 8 & 20 & 195 & 5.0 & 5 & 14 & 81 & 7.1 & 6 & 13 & 154 & 30.8 \\
2 & 2 & 6 & 26 & 50.0 & 2 & 5 & 72 & 40.0 & 5 & 14 & 162 & 35.7 \\
3 & 4 & 13 & 105 & 23.1 & 6 & 17 & 224 & 35.3 & 5 & 14 & 119 & 42.9 \\
4 & 5 & 16 & 132 & 18.8 & 6 & 20 & 127 & 20.0 & 4 & 12 & 85 & 41.7 \\
\hline
Aggregated & 19 & 55 & 458 & 24.2 & 19 & 56 & 504 & 25.6 & 20 & 53 & 520 & 37.8 \\
    \end{tabular}
    }
    \caption{Annotations. We report the statistics for the number instances annotated, the amount of evidence snippets annotated, the total number of reports in the annotated instances, and the percent of evidence annotated as ``Useful'' and ``Very Useful''. Aggregated statistics are computed by summing over the annotators except in the case of ``Percent Useful'', where scores are macro-averaged over annotators. (This is slightly different from Figure \ref{fig:model-evidence} where percentages are macro-averaged, i.e., we combine all annotated evidence).}
    \label{tab:annotations}
\end{table*}

\begin{figure}
    \centering
    \includegraphics[scale=.65]{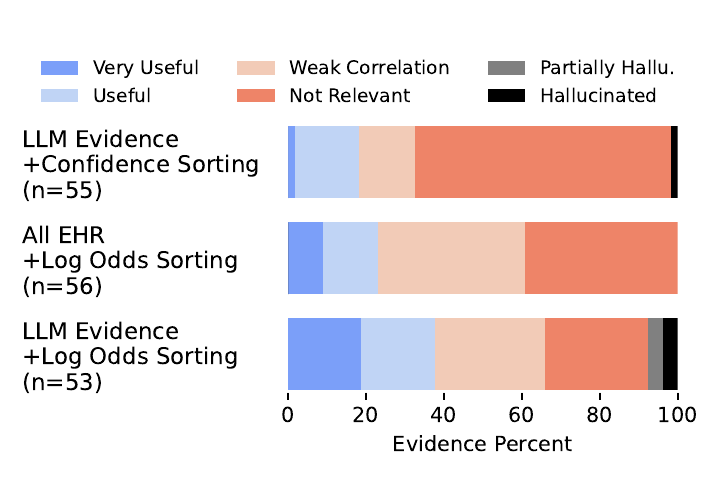}
    \caption{\textbf{Evidence Usefulness} (the maximum score across conditions) for our approach and two ablations. ``LLM Evidence+Confidence Sorting'' uses model evidence, but sorts by (length-normalized) log probability instead of the log odds. ``All EHR+Log Odds Sorting'' does not use LLM evidence and instead takes the last 1000 sentences in the record as evidence.}
    \label{fig:model-evidence}
\end{figure}

The majority of our results are based on annotations from 4 annotators on 24 instances and 3 models.
Each instance has a maximum of 3 annotators, each annotating different models (assigned randomly). 
Table \ref{tab:annotations} reports detailed statistics. 

Our main goal is to understand if our approach can retrieve better evidence.
To this end, we plot the percentage of evidence annotated in each category of usefulness for each model in Figure \ref{fig:model-evidence}.\footnote{Sometimes annotators noticed nearly duplicate evidence, so we kept track of this evidence (a total of 21 snippets) and omitted it from the results.}
Though we record usefulness for each condition individually, here we combine these annotations by taking the maximum score across the conditions for each piece of evidence.
To identify hallucinated evidence, we conducted post-hoc annotations with only the annotated LLM-generated evidence that was abstractive (42 of 108).\footnote{To annotate hallucinations, we provided a clinician the generated evidence alongside the report from which it was generated and asked if the evidence was hallucinated or partially hallucinated.
Full results are in appendix Table \ref{tab:hallucinations}.}
The results highlight the necessity of both the ``\textbf{LLM Evidence}'' retrieval component and the ``\textbf{Log Odds Sorting}'' method, as both other variants retrieve significantly less ``Useful'' and ``Very Useful'' evidence and more ``Weakly Correlated'' and ``Not Relevant'' evidence.
We also find a relatively small number of hallucinations (5) and note where the hallucinated evidence was originally ranked in Table \ref{tab:hallucination_ratings}.

\begin{figure}
    \centering
    \includegraphics[scale=.52]{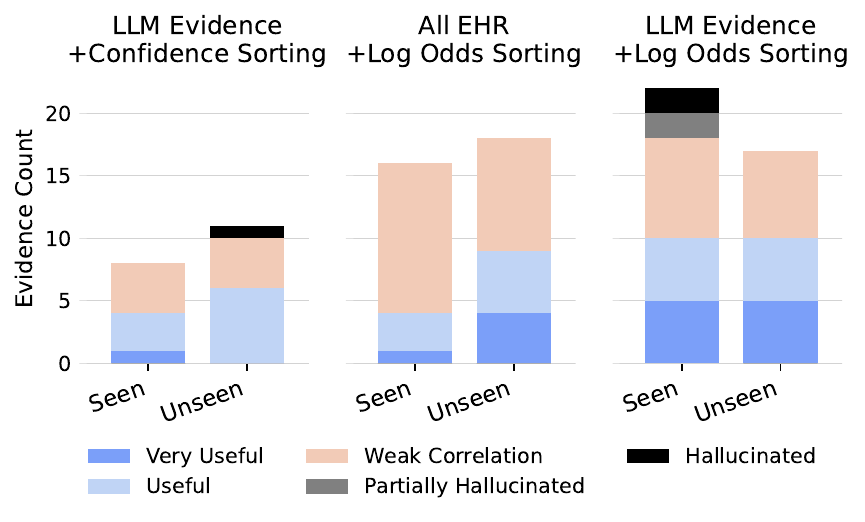}
    \caption{\textbf{Seen vs. unseen} evidence counts for all evidence that at least weakly correlates with a condition. Curiously, the LLM Evidence with Log Odds Sorting model has some hallucinated evidence that was seen by annotators. See section \ref{section:results} for a discussion.}
    \label{fig:seen-vs-unseen}
\end{figure}

\begin{table}[]
    \footnotesize
    \centering
\begin{tabular}{c c c c c}
& Not & Weakly & & Very \\
& Relevant & Correlated & Useful & Useful \\
\hline
LLM Conf. & 0 & 1 & 0 & 0 \\
Log Odds & 0 & 1 & 3 & 0 \\
\end{tabular}
    \caption{The original evidence ratings of hallucinations.}
    \label{tab:hallucination_ratings}
\end{table}

How much of the relevant retrieved evidence is redundant with the information already uncovered during the annotator's initial review of the patient?
We plot evidence counts separately for seen vs unseen evidence in Figure \ref{fig:seen-vs-unseen} and find that there is a significant amount of unseen evidence that is useful and very useful in all models.
It is interesting to note that some hallucinated evidence was ``seen'' by annotators.
We believe this is most likely due to some hallucinated evidence having been potentially true of the patient at some point but not with respect to the specific report used to generate it (e.g. the generated evidence says the patient has a bleeding colon lesion, but the report says that the patient \textit{no longer} has this; see Table \ref{tab:hallucinations} for more examples).

The rated usefulness of evidence does not necessarily matter if it does not 
affect the clinician's decision.
An example of how these models might work in practice is when our LLM Evidence model with Confidence Sorting surfaced the following: ``Atrial fibrillation with rapid ventricular response. Compared to the previous tracing atrial fibrillation is seen. Other findings are similar. The patient is at risk of pulmonary edema.''
In this case the annotator changed their estimate of the likelihood of pulmonary edema from unlikely to somewhat likely, and it turns out that pulmonary edema did appear in a future report. 

We show all 7 instances where annotators changed their mind after viewing evidence in Appendix Table \ref{tab:evidence-that-impacts-annotator}. 
Of these we find 2 instances (including the example above) where annotators' increased their likelihood of conditions that were extracted from future records, and 5 where condition(s) other than the synthetically labeled condition(s) were affected (mostly by increasing the annotators' risk assessments).
Though more data should be collected, this indicates the model might improve annotator recall (though at some cost in precision); recall is arguably more important here. 

\begin{figure}
    \centering
    \includegraphics[scale=.67]{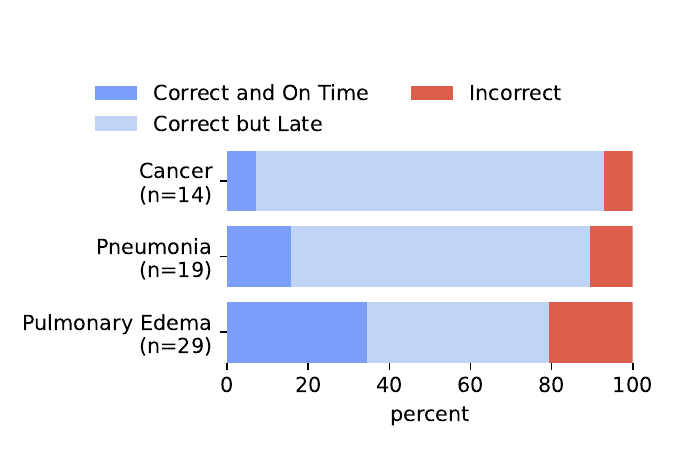}
    \caption{Synthetic label precision. For each confident diagnosis label extracted by the system, annotators check whether the diagnosis actually appears in the report (and is definitive), and subsequently if subjectively they believe that report is likely the \textit{first} time the diagnosis was definitive based on the report language.}
    \label{fig:synthetic-label-precision}
\end{figure}

Given that we are using synthetic labels of future diagnoses for both training and evaluation for risk prediction (discussed next), it is important to evaluate how well our labels align with ground truth.
Given that ICD codes are not fine-grained enough and are not always accurate, we turn to manual annotations of precision for this evaluation.
In Figure \ref{fig:synthetic-label-precision}, we report the precision of these labels for being correct or for being ``correct and on time''. This second category is a stronger correctness in which the annotator also noted that the note where the label was detected subjectively seems to be the first note where that label should have been given as judged using the phrasing in the note.\footnote{It would be time-consuming to annotate this directly becuase it involves looking at a lot of prior notes.}

We see reasonable precision when using automatic labeling with the LLM pipeline (about 80 percent and above for all conditions).
We also compute inter-annotator agreement for these annotations of precision across the 4 annotators by enforcing that 8 annotated predictions overlap for all the annotators.
The Fleiss' Kappa score for these synthetic label annotations was .68 for the 3-category classification shown in Figure \ref{fig:synthetic-label-precision} and .86 for the 2-category classification obtained by simplifying the labels into just  ``Correct'' or ``Incorrect''.

\begin{figure}
    \centering
    \includegraphics[scale=.63]{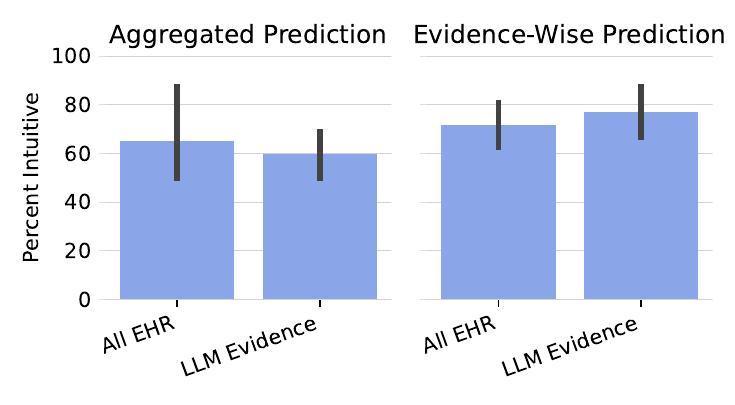}
    \caption{Intuitiveness of predictions macro-averaged across annotators.}
    \label{fig:prediction-intuitiveness}
\end{figure}

We would also like to assess how well our models' risk estimates aligns with the intuitions of clinicians with respect to the aggregated and individual predictions.
Though for the aggregated prediction for an instance, we ask annotators to take the magnitude of the risk, not just the direction (i.e. increased compared to baseline or decreased compared to baseline) into account, for evidence-level predictions, we ask annotators to take the magnitude with a grain of salt and mostly judge based on the direction.
This is because the magnitudes appeared to be somewhat artificially inflated potentially either due to the strong evidence trying to ``compensate'' for the evidence that does not actively contribute to the log odds (see Figure \ref{fig:evidence-log-odds-hist}) or because of the sorting method.\footnote{Future work might investigate how to bring make this \textit{magnitude} more interpretable.}
Figure \ref{fig:prediction-intuitiveness} shows that both models do reasonably well with respect to the aggregated and evidence-wise predictions, and both do slightly better on evidence-wise as opposed to aggregated predictions.

\begin{figure}
    \centering
    \includegraphics[scale=.75]{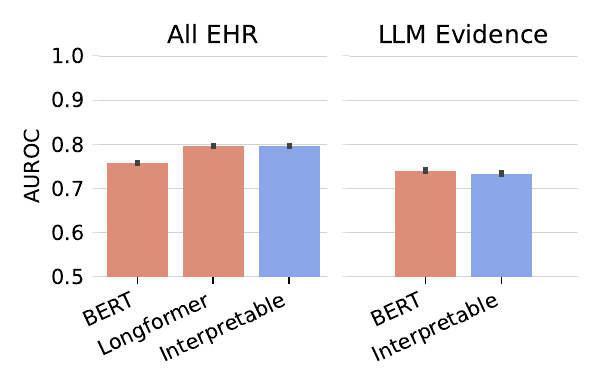}
    \caption{Macro-averaged risk prediction performance evaluated on synthetic labels and averaged over 5 random seeds for choosing the which time-point in the EHR to use prior to the diagnosis label.
    Error bars represent the standard deviation of the random seeds.
    Here, BERT and Longformer refer to Clinical BERT and Clinical Longformer.}
    \label{fig:rp-performance}
\end{figure}

Finally, it is important to evaluate the actual prediction performance of our models on our synthetic labels.
Here we also compare against baseline models that are not interpretable: BERT and Longformer.
These black-box models are trained on both the All EHR and the concatenated retrieved LLM evidence.
Figure \ref{fig:rp-performance} shows that including all evidence usually helps prediction performance, but using the blackbox vs interpretable models on the same input does not effect performance.

\section{Conclusions}

Clinicians should have access to all the pertinent information to make well-grounded decisions for diagnosing a patient, but currently they are inundated with (unstructured) information from the EHR. 
This is exacerbated by the time constraints faced by practitioners.
We have proposed an approach that aims to facilitate efficient access to potentially important data within EHR; our method capitalizes on the capabilities of LLMs to produce digestable, abstractively generated text evidence, which is then consumed by a Neural Additive Model (NAM) to yield a prediction.

We find that using NAMs does not sacrifice predictive quality, but does enable models to surface useful evidence to clinicians.
Using the LLM to create the starting set of evidence to feed into the NAM does sacrifice some performance, but it also significantly increases the usefulness of the evidence in comparison with using the raw sentences from EHR notes as evidence.

Further, we find that in some cases the surfaced evidence is able to change a clinician's mind, increasing the clinician's recall though decreasing precision, which warrants future work to improve on this system.
One major concern is that this type of system could increase clinician's workload rather than decrease it.
Future work should assess exactly how and when it might be beneficial to show snippets to clinicians.

\section{Limitations}

The proposed approach of combining abstractive LLM evidence with Neural Additive Models shows promise, but there are still many concerns that need to be addressed in future work.
One of the biggest concerns is about the use of abstractive ``evidence'' produced by LLMs.
Though our analysis does not find many hallucinations, their existence certainly poses risks and should be studied further in future work.
Any hallucinated evidence could at best negatively impact trust of clinicians in the system and at worst mislead clinicians and negatively affect patient outcomes.
We also did not experiment much with different prompts or models for producing this evidence given that our main focus was on validating the system-level approach rather than individual components.

Another limitation concerns the lack of a significant number of baseline models.
Though not many baselines exist for a task that involves retrieving evidence supporting predictions in EHR, there are still potential baselines that use relevance weights or cosine similarity with clinical BERT that we could have included.
However, due to the extensive amount of time needed for just one annotation on one model, we chose to focus on ablating over the LLM evidence retrieval and sorting method components of the model.

Finally, our analysis mostly relies on a relatively small amount of annotations from one dataset.
This again stems from the time cost of annotations.
Each annotator must first look through a whole patient's record to get a sense of the patient before even getting to any annotations.
On average, this took almost 3 minutes, which is all before annotators even see any of the questions.
Then, because the study focuses on just the top evidence presented for each instance, each annotator only annotates 3.2 evidence snippets on average per instance.
This time-consuming process did limit the number of annotations we could obtain.

\section*{Acknowledgements}

This work was supported in part by the National Science Foundation (NSF/IIS-1901117), and by the the National Institutes of Health (NIH) under the National Library of Medicine (NLM; 1R01LM013772).

\bibliography{anthology,custom}

\appendix
\onecolumn
\newpage
\twocolumn
\section{Dataset and {Preprocessing}}\label{sec:dataset-and-preprocessing}

We treat each patient as an instance and split the instances randomly into a training split for training the risk prediction model, a validation split for picking the best checkpoint and other hyperparmeter tuning, a test split for automatically evaluating the risk prediction, and an annotation split for annotations.
After the first round of annotations, because we changed our model (see section \ref{sec:prompting-problems}), we throw out all patients annotated in the first round so that the second and final round of annotations, which were used to compute all results, were conducted on a held-out set of instances.
Instance order was randomized, so no bias resulted from throwing out the first set of instances annotated.

Each instance is randomly separated into a past and future.
During training, repeated examples might have different samples time-points, but during evaluation and annotation, the same randomly-picked time-point is used across all evaluations and annotations.
We also ignore examples longer than 200 reports for computational purposes.
Given that this application's use case is for lengthy records, for annotations we restricted to instances with greater than 10 records for all but 3 annotated instances, which had already been completed.

During training, to overcome problems caused by data imbalance and for computational reasons, we randomly sub-sample 20\% of the negative examples---i.e., examples that have none of the three considered conditions.
For annotations, we sub-sample negatives such that each annotation has a 50\% chance of having at least one positive condition of the three considered.

\section{Evidence Retrieval Details}\label{sec:evidence-prompt}

We use the same prompts as in \cite{ahsan2023retrieving} for retrieving evidence of risks and signs.
We also add an additional set of two prompts for retrieving evidence relating to a particular queried risk factor.
The exact prompts used are as follows:

\noindent
\textbf{Evidence of Risk}

\noindent
Prompt 1:
\begin{quote}
Read the following clinical note of a patient:\\
\texttt{<input>}\\
Question: Is the patient at risk of <query>? Choice: -Yes -No\\
Answer: 
\end{quote}
\vspace{20pt}
\noindent
Prompt 2:
\begin{quote}
Read the following clinical note of a patient:\\
\texttt{<input>}\\
Based on the note, why is the patient at risk of <query>?\\
Answer step by step:
\end{quote}

\noindent
\textbf{Evidence of Signs}

\noindent
Prompt 1:
\begin{quote}
Read the following clinical note of a patient:\\
\texttt{<input>}\\
Question: Does the patient have <query>? Choice: -Yes -No\\
Answer: 
\end{quote}

\noindent
Prompt 2:
\begin{quote}
Read the following clinical note of a patient:\\
\texttt{<input>}\\
Question: Extract signs of <query> from the note.\\
Answer: 
\end{quote}

\noindent
\textbf{Evidence of a Queried Risk Factor}

\noindent
Prompt 1:
\begin{quote}
Read the following clinical note of a patient:\\
\texttt{<input>}\\
Question: Does the patient have <query>? Choice: -Yes -No\\
Answer: 
\end{quote}

\noindent
Prompt 2:
\begin{quote}
Read the following clinical note of a patient:\\
\texttt{<input>}\\
What evidence is there that the patient has <query>?\\
Answer: 
\end{quote}

\section{Risk Prediction Inputs}\label{sec:risk-prediction-inputs}

To provide some context of the evidence for the risk prediction model, we decided to add some metadata to the evidence when it was presented to the model with the hope that the model could use this context to make better predictions.
In particular, we decided to include the query that was used to retrieve a piece of evidence, and the relative day of the report from which the evidence was retrieved in the following format:
\begin{quote}
    <query> (<query\_type>): ``<evidence>'' (day <relative\_day>)
\end{quote}
For example, if querying a diagnosis of ``pneumonia'' retrieved the evidence ``the patient has a cough.'' from a report 5 days prior to the current time-step, the evidence would be be presented to the model as:
\begin{quote}
    pneumonia (diagnosis): ``the patient has a cough.'' (day -5)
\end{quote}

\section{Certain Diagnosis Extraction Prompts}\label{sec:full-prompts}

\textbf{Prompt 1}:
\begin{quote}
Read the following report:

\texttt{<input>}

Question: Is there a confident diagnosis of the patient's condition? Choice: -Yes -No\\
Answer: 
\end{quote}

\noindent
\textbf{Prompt 2}:
\begin{quote}
Read the following report:

<input>

Answer step by step: What is the correct diagnosis of the patient's condition?\\
Answer:
\end{quote}
We use Chain of Thought (CoT) prompting here because---similar to the evidence retrieval step---we want the model first to extract the parts of the report that refer to a diagnosis, as this seems to work better than going straight to the list of diagnoses.
In initial experiments, using the CoT prompt appeared to more easily elicit these verbose extractions.

\noindent
\textbf{Prompt 3}:

\begin{quote}
Here is a diagnosis of a patient:

<confident diagnosis>

Question: Provide a list of diagnostic terms or write none.\\
Answer: 
\end{quote}

\section{Prompting Problems}\label{sec:prompting-problems}

\begin{figure}
    \centering
    \includegraphics[scale=.7]{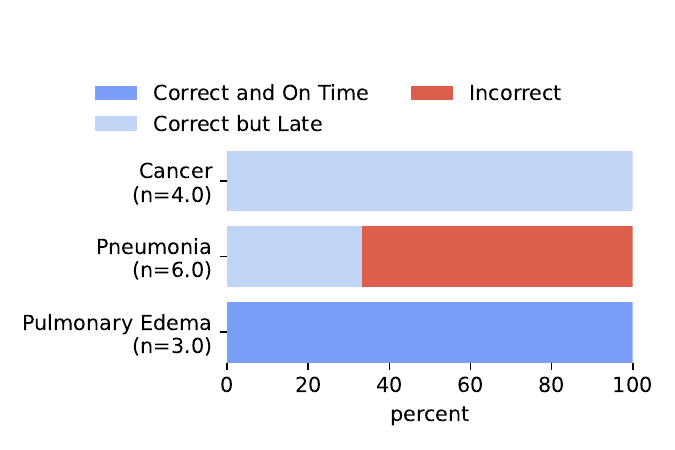}
    \caption{Synthetic labels on validation examples before correcting the prompting problem.}
    \label{fig:synthetic-labels-before-correcting-prompting-problems}
\end{figure}

In our 3-stage prompting process, we initially had some problems with false positives in scenarios where pneumonia was negated (Figure \ref{fig:synthetic-labels-before-correcting-prompting-problems}).
We discovered that this was because our 3rd prompt was originally:
\begin{quote}
Here is a diagnosis of a patient:

<confident diagnosis>

Question: Based on this diagnosis, provide a list of diagnostic terms.\\
Answer: 
\end{quote}
This particular prompt sometimes produced positive synthetic labels for pneumonia when pneumonia was actually negated in the confident diagnosis generated by the previous prompt.
We realized this when starting to annotate validation examples, so we changed our prompt (see section \ref{sec:synthetic-label-extraction}).

We also noticed that some false positives might be caused by the model treating the admitting diagnosis as true, even though it can often be wrong according to the report text.
To combat this, we added a preprocessing step before inserting the report into the confident diagnosis extraction prompts that removed the admitting diagnosis from the text.
All of the test annotations used for the results do not include or overlap patients with the annotated examples which were used in this phase (chosen from the randomly shuffled annotation split) and precipitated these modifications.

\section{Description of Terms for Models and Settings}\label{sec:terms}

Table \ref{tab:terms} shows all of the terms used to describe different models and settings.

\begin{table*}[]
    \centering
    \footnotesize
    \begin{tabular}{l p{12cm}}
        \hline
        LLM Evidence & Models that use the evidence retrieved with an LLM. \\
        All EHR & Models that use the all of the text in the EHR. For Interpretable Neural Additive Model, this text is split at the sentence level. \\
        BERT or Longformer & Blackbox models that take either All EHR or LLM Evidence (concatenated) as input. BERT refers to Clinical BERT \cite{alsentzer-etal-2019-publicly} and Longformer refers to Clinical-Longformer \cite{li2023comparative}. \\
        Interpretable & The proposed Interpretable Neural Additive Model, which can operate either on LLM Evidence or All EHR inputs. \\
        Confidence Sorting & Sorting LLM Evidence by the length-normalized log-likelihood of the evidence under the LLM. \\
        Log Odds Sorting & Sorting either LLM Evidence or All EHR inputs by the mean squared error of the predicted log odds (equation \ref{eq:mse}). \\
        \hline
    \end{tabular}
    \caption{Description of terms.}
    \label{tab:terms}
\end{table*}

\section{Experiments}

We use Clinical BERT for the NAM prediction model.
For all models, we train for up to 10 epochs on one Quadro RTX 8000 GPU and pick the best checkpoint (where checkpoints occur every 5 percent of an epoch).
For the LLM for both evidence retrieval and synthetic label extraction we use FLAN-T5-XXL \citep{chung2022scaling, 51119}.
In the case of All EHR used as input to the interpretable NAM, we split sentences with NLTK.

\section{Usefulness of Queries}

\begin{figure*}
    \centering
    \includegraphics[scale=.8]{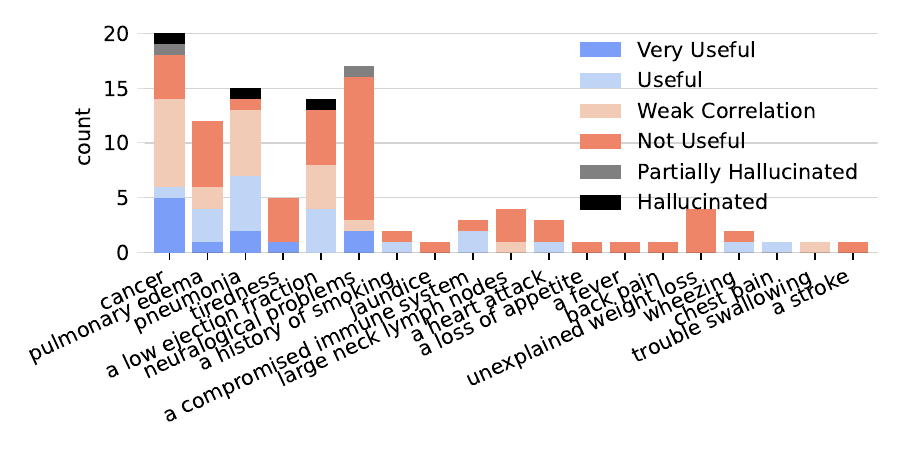}
    \caption{Usefulness per Query.}
    \label{fig:query-usefulness}
\end{figure*}

\begin{table*}[]
    \centering
    \begin{tabular}{c p{12cm}}
Diagnosis & Risk Factors \\
\hline
Pneumonia & a stroke, trouble swallowing, a compromised immune system, a high white blood cell count, a fever \\
Pulmonary Edema & a low ejection fraction, a heart attack, steroid use \\
Cancer & back pain, neuralogical problems, a history of smoking, night sweats, unexplained weight loss, a chronic cough with blood, large neck lymph nodes, a loss of appetite, jaundice, chest pain, hoarseness, tiredness, wheezing
    \end{tabular}
    \caption{A non-exhaustive list of risk factors proposed by a clinician for use in queries.}
    \label{tab:risk-factors}
\end{table*}

Unlike \cite{ahsan2023retrieving}, we do not directly evaluate how relevant the retrieved evidence is to the query used to retrieve it; we instead focus on how relevant the evidence is to the risk predictions.
However, we would like to examine which queries produce useful evidence.
Figure \ref{fig:query-usefulness} shows counts of evidence in each category separated across which query was used to retrieve that evidence.
It seems as though the most useful evidence came from the three queries that directly ask about the condition for which we are predicting risk (the three left-most queries), but a few additional queries sometimes did prove useful.

\section{Full Prediction Performance}
\label{sec:full-prediction-performance}

We report the full prediction performance in Table \ref{tab:full-prediction-performance}.

\begin{table*}[]
    \centering
    \footnotesize
    \begin{tabular}{l|c c c c}
& AUROC & Precision & Recall & F1 \\
\hline
BERT (All EHR) & 75.6  $\pm$ .19 & 65.6  $\pm$ 1.38 & 16.8  $\pm$ .38 & 26.8  $\pm$ .43 \\
Longformer (All EHR) & 79.6  $\pm$ .22 & 55.5  $\pm$ .32 & 28.8  $\pm$ .43 & 37.9  $\pm$ .38 \\
Interpretable (All EHR) & 79.5  $\pm$ .23 & 56.5  $\pm$ .57 & 20.5  $\pm$ .58 & 30.1  $\pm$ .60 \\
BERT (LLM Evidence) & 74.0  $\pm$ .27 & 51.6  $\pm$ 1.32 & 22.7  $\pm$ .27 & 31.5  $\pm$ .42 \\
Interpretable (LLM Evidence) & 73.3  $\pm$ .27 & 53.6  $\pm$ 1.09 & 15.0  $\pm$ .36 & 23.4  $\pm$ .48 \\
    \end{tabular}
    \caption{Macro-averaged \textbf{risk prediction performance} on the synthetic labels averaged over 5 different random seeds used for choosing the time-point in each patient that separates the past from the future.}
    \label{tab:full-prediction-performance}
\end{table*}

\section{Annotators Changing Their Minds}

Table \ref{tab:evidence-that-impacts-annotator} presents all the occurrences of annotators changing their mind.

\begin{table*}[]
    \centering
    \resizebox{\textwidth}{!}{
    \begin{tabular}{c c c p{2cm} p{6cm} p{4cm} p{2cm}}
Annotator & Model & Sorting & Changes & Best Evidence & Usefulness & Synthetic Label \\
\hline
2 & LLM Evidence & Confidence Sorting & Pneumonia: Unlikely $\rightarrow$ Somewhat likely & There is a small right pneumothorax. There is extensive consolidation of the right upper lobe. Consolidation in the right lower lobe is mostly located in the superior segment. The left lung is grossly clear. There. Signs: There is extensive consolidation of the right upper lobe. Consolidation in the right lower lobe is mostly located in the superior segment. The left lung is grossly clear. There is no left pleural effusion. There is & Useful for \textbf{Pneumonia} & Pneumonia \\
4 & LLM Evidence & Confidence Sorting & Pulmonary Edema: Unlikely $\rightarrow$ Somewhat likely & Atrial fibrillation with rapid ventricular response. Compared to the previous tracing atrial fibrillation is seen. Other findings are similar. The patient is at risk of pulmonary edema. & Useful for \textbf{Pulmonary Edema} & Pulmonary Edema \\
3 & All EHR & Log Odds Sorting & Cancer: Unlikely $\rightarrow$ Very likely & Basal cell skin ca. [**27**]. & Useful for \textbf{Cancer} & Pulmonary Edema \\
4 & All EHR & Log Odds Sorting & Cancer: Unlikely $\rightarrow$ Somewhat likely & o.b.resident to see pt., pt.waiting for a "biopsy". & Useful for \textbf{Cancer} & Pulmonary Edema \\
4 & All EHR & Log Odds Sorting & Pulmonary Edema: Somewhat likely $\rightarrow$ Unlikely, Pneumonia: Somewhat likely $\rightarrow$ Very likely & There is increased opacity in the.  retrocardiac left lower lobe, as well as the right lower lobe, which could be.  due to atelectasis, aspiration, or possibly pneumonia. & Very Useful for \textbf{Pneumonia} &  \\
1 & LLM Evidence & Log Odds Sorting & Pneumonia: Somewhat likely $\rightarrow$ Very likely & CXR showed L middle/lower lobe PNA, prob asp PNA. & Very Useful for \textbf{Pneumonia} &  \\
4 & LLM Evidence & Log Odds Sorting & Cancer: Unlikely $\rightarrow$ Very likely & CLL. Signs: id: pmh of CLL & Very Useful for \textbf{Cancer} &  \\
    \end{tabular}
    }
    \caption{Examples of the 5 instances where annotators changed their mind based on evidence shown.}
    \label{tab:evidence-that-impacts-annotator}
\end{table*}


\section{Ablation over amount of evidence used}

\begin{figure}
    \centering
    \includegraphics[scale=.8]{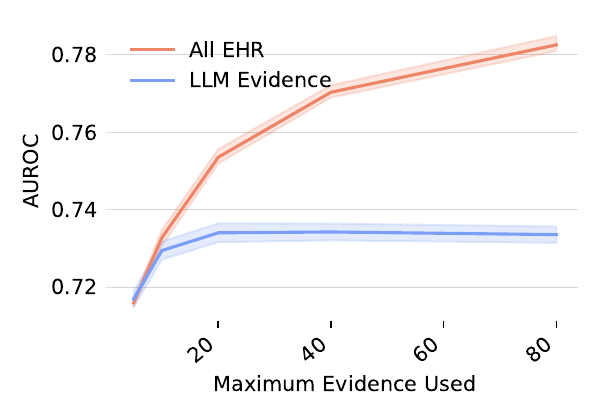}
    \caption{Ablation over amount of evidence used to make a risk prediction.}
    \label{fig:rp-evidence-ablation}
\end{figure}

Figure \ref{fig:rp-evidence-ablation} shows performance if we limit to a set amount of evidence that can be used in the Neural Additive Model's final aggregated score.
This shows that the model performance is not affected until it is limited to using less than 20 snippets for predictions.

\section{Evidence Histograms}

\begin{figure}
    \centering
    \includegraphics[scale=.6]{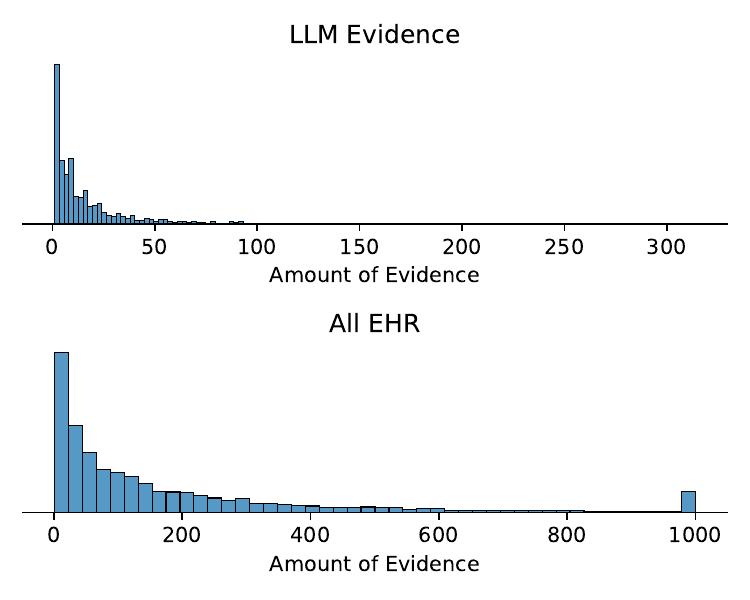}
    \caption{Histogram of the number of text snippets for each instance.}
    \label{fig:amount-of-evidence-hist}
\end{figure}

\begin{figure}
    \centering
    \includegraphics[scale=.6]{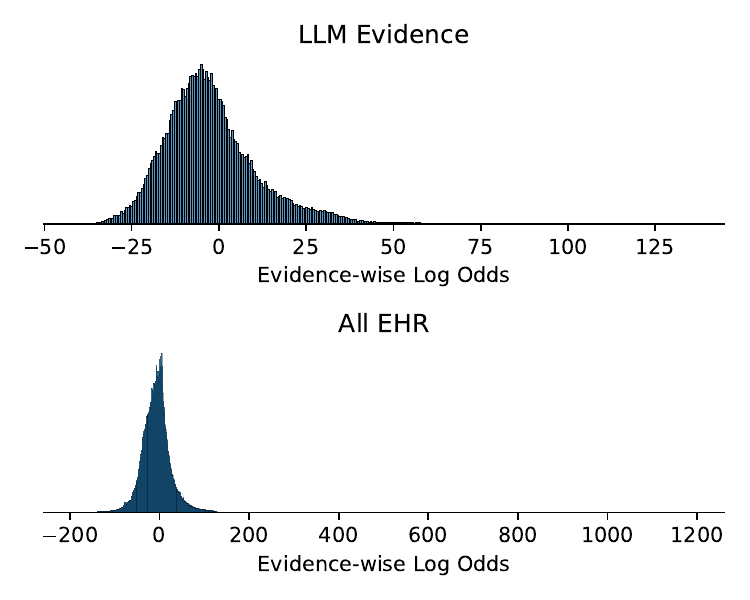}
    \caption{Histogram of the log odds of each individual piece of evidence.}
    \label{fig:evidence-log-odds-hist}
\end{figure}

Figure \ref{fig:amount-of-evidence-hist} shows a histogram of the amount of evidence per each instance, and Figure \ref{fig:evidence-log-odds-hist} shows what the distribution over the log odds votes looks like.

\section{Annotation Interface}

\begin{figure*}
    \centering
    \includegraphics[scale=.38]{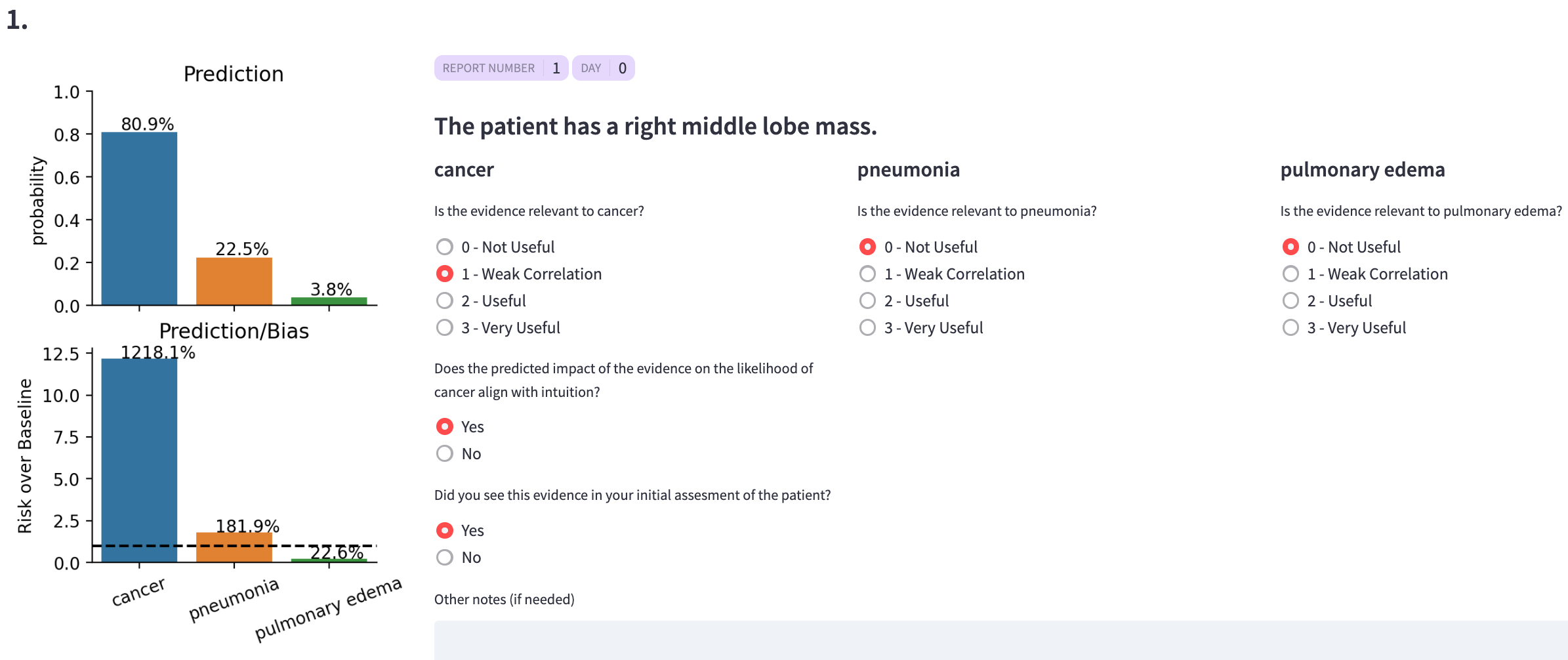}
    \caption{An example part of the \textbf{evidence} annotation interface. The plots on the left indicate the predicted likelihood (top) and the odds ratio (bottom).}
    \label{fig:evidence_ann_interface}
\end{figure*}

Figure \ref{fig:evidence_ann_interface} shows a screenshot of what the part of the interface dedicated to annotating evidence looks like.

\section{Hallucinations}

Table \ref{tab:hallucinations} shows all of the annotated evidence that was subsequently marked as a hallucination along with an explanation of why it is a hallucination and other information about the evidence.

\begin{table*}[]
    \footnotesize
    \centering
\begin{tabular}{p{2.7cm} p{1.4cm} p{3.5cm} p{1.5cm} p{1.5cm} p{1cm} p{1.5cm}}
Evidence & Hallucination & Explanation & Query & Sorting & Seen & Rating \\
\hline
The patient has a bleeding colon lesion. & Yes & The report indicates that the patient used to have a bleeding colon lesion but no longer does. & cancer & Log Odds Sorting & Yes & Useful \\
The patient has a history of heart failure. & Yes & The report looks like it is cut off, and the only thing mentioned is a Coronary artery bypass graft (CABG). & a low ejection fraction & Log Odds Sorting & Yes & Useful \\
The patient has a history of sepsis. & Yes & Report says ``R/O'' meaning rule out sepsis. & pneumonia & LLM Confidence & No & Weakly Correlated \\
The patient has a mass in her breast. & Partially & The report header says that the patient has a mass, but the body of the report does not indicate this. & cancer & Log Odds Sorting & Yes & Weakly Correlated \\
The patient had a brain tumor removed. & Partially & Clinicians do not usually refer to pituitary adenomas (which the report indicates) as brain tumors. & neuralogical problems & Log Odds Sorting & Yes & Useful \\
\end{tabular}
    \caption{Clinician-annotated hallucinations.}
    \label{tab:hallucinations}
\end{table*}

\end{document}